\documentclass[11pt,a4paper]{article}
\usepackage[utf8]{inputenc}
\usepackage{hyperref}  % needed for cleveref on arXiv
\usepackage{amsmath}
\usepackage[capitalize,nameinlink]{cleveref}
\usepackage[hyperref]{acl2018_pagenums}
\usepackage{times}
\usepackage{latexsym}
\usepackage{amssymb}
\usepackage{anyfontsize}
\usepackage{array}
\usepackage[inline]{enumitem}
\usepackage{graphicx}
\usepackage{multirow}
\usepackage{pifont}
\usepackage[normalem]{ulem} % sout ... strikeout
\usepackage{calc}
\usepackage{tabularx}
\usepackage{url}

\aclfinalcopy % Uncomment this line for the final submission
\def\equo#1{``#1''}
\def\footurl#1{\footnote{\url{#1}}}
\def\clap#1{\hbox to 0pt{\hss #1\hss}}

\def\Tref#1{\cref{#1}}

\def\subtref#1{\cref{#1} in supplementary material}
\def\textsubscript#1{$_\text{#1}$}

\def\perscite#1{\citet{#1}}
\def\parcite#1{\cite{#1}}

\DeclareMathOperator*{\softmax}{softmax}

\newcolumntype{L}[1]{>{\raggedright\let\newline\\\arraybackslash\hspace{0pt}}m{#1}}
\newcolumntype{C}[1]{>{\centering\let\newline\\\arraybackslash\hspace{0pt}}m{#1}}
\newcolumntype{R}[1]{>{\raggedleft\let\newline\\\arraybackslash\hspace{0pt}}m{#1}}

\newcommand{\cmark}{\ding{51}}
\newcommand{\xmark}{\ding{55}}

\newcommand{\parens}[1]{\llap{(}#1\rlap{)}}
\newcommand{\tbf}{\fontseries{b}\selectfont}

\def\ourtitle{Are BLEU and Meaning Representation in Opposition?}
\title{\ourtitle}

\author{Ond\v rej C\' ifka \and Ond\v rej Bojar \\
  Charles University \\
	Faculty of Mathematics and Physics \\
	Institute of Formal and Applied Linguistics \\
  \texttt{\{cifka,bojar\}@ufal.mff.cuni.cz}}

\date{}

\begin{document}
\maketitle
\begin{abstract}
One of possible ways of obtaining con\-tin\-u\-ous-space sentence
representations is by training neural machine translation (NMT) systems. The
recent attention mechanism however removes the single point in the neural network
from which the source sentence representation can be extracted. We propose several
variations of the attentive NMT architecture bringing this meeting point back. Empirical evaluation suggests that the
better the translation quality, the worse the learned sentence representations
serve in a wide range of classification and similarity tasks.
\end{abstract}

\def\BLEUmmm{%
de-\textsc{attn} & --- & --- & 37.6 & 36.2 \\
de-\textsc{trf} & --- & --- & 38.2 & 36.1 \\
de-\textsc{attn-attn} & 2400 & 12 & 36.2 & 34.8 \\
de-\textsc{attn-attn} & 1200 & 12 & 35.6 & 34.3 \\
de-\textsc{attn-attn} & 600 & 8 & 35.4 & 33.7 \\
de-\textsc{attn-attn} & 600 & 12 & 35.3 & 33.4 \\
de-\textsc{attn-attn} & 1200 & 6 & 35.0 & 33.2 \\
de-\textsc{attn-attn} & 600 & 6 & 35.1 & 33.2 \\
de-\textsc{trf-attn-attn} & 600 & 3 & 32.3 & 30.1 \\
de-\textsc{attn-attn} & 600 & 3 & 31.4 & 29.4 \\
de-\textsc{attn-ctx} & 1200 & 12 & 30.6 & 29.2 \\
de-\textsc{attn-ctx} & 600 & 12 & 29.8 & 29.1 \\
de-\textsc{attn-ctx} & 600 & 8 & 29.8 & 28.9 \\
de-\textsc{attn-ctx} & 600 & 6 & 29.5 & 28.8 \\
de-\textsc{trf-attn-attn} & 2400 & 12 & 30.6 & 28.5 \\
de-\textsc{maxpool-ctx} & 600 & --- & 27.8 & 28.1 \\
de-\textsc{final-ctx} & 600 & --- & 28.1 & 26.9 \\
de-\textsc{attn-ctx} & 600 & 3 & 27.8 & 26.9 \\
de-\textsc{avgpool-ctx} & 600 & --- & 27.1 & 26.5 \\
de-\textsc{attn-attn} & 600 & 1 & 27.2 & 26.0 \\
de-\textsc{trf-attn-attn} & 600 & 6 & 26.5 & 25.8 \\
de-\textsc{trf-attn-attn} & 1200 & 12 & 26.6 & 25.3 \\
de-\textsc{final} & 600 & --- & 23.9 & 23.8 \\
}%
\def\BLEUcze{%
cs-\textsc{attn} & --- & --- & 22.8 & 22.2 \\
cs-\textsc{attn-attn} & 1000 & 8 & 19.1 & 18.4 \\
cs-\textsc{attn-attn} & 4000 & 4 & 18.4 & 17.9 \\
cs-\textsc{attn-attn} & 1000 & 4 & 17.5 & 17.1 \\
cs-\textsc{attn-ctx} & 1000 & 4 & 16.6 & 16.1 \\
cs-\textsc{final-ctx} & 1000 & --- & 16.1 & 15.5 \\
cs-\textsc{attn-attn} & 1000 & 1 & 15.3 & 14.8 \\
cs-\textsc{final} & 1000 & --- & 11.2 & 10.8 \\
cs-\textsc{avgpool} & 1000 & --- & 11.1 & 10.6 \\
cs-\textsc{maxpool} & 1000 & --- & 5.4 & 5.4 \\
}%
\def\SentEvalA{%
InferSent & 4096 & --- & 81.5 & 86.7 & 92.7 & 90.6 & 85.0 & 45.8 & 88.2 & 76.6 & 86.4 & 83.7 & 81.7 \\
GloVe-BOW & 300 & --- & 77.0 & 78.2 & 91.1 & 87.9 & 81.0 & 44.4 & 82.0 & 72.3 & 78.2 & 66.0 & 75.8 \\
cs-\textsc{final-ctx} & 1000 & --- & 68.7 & 77.4 & 88.5 & 85.5 & 73.0 & 38.2 & 88.6 & 71.8 & 82.1 & 70.2 & 74.4 \\
cs-\textsc{attn-attn} & 1000 & 1 & 68.2 & 76.0 & 86.9 & 84.9 & 72.0 & 35.7 & 89.0 & 70.7 & 80.8 & 69.3 & 73.4 \\
cs-\textsc{final} & 1000 & --- & 67.9 & 75.7 & 87.6 & 84.7 & 72.5 & 36.2 & 86.0 & 71.4 & 81.1 & 69.2 & 73.2 \\
cs-\textsc{maxpool} & 1000 & --- & 67.4 & 75.2 & 86.9 & 84.3 & 70.3 & 37.5 & 85.8 & 72.1 & 81.7 & 68.5 & 73.0 \\
cs-\textsc{avgpool} & 1000 & --- & 66.5 & 74.1 & 86.5 & 85.0 & 71.9 & 36.7 & 85.4 & 70.0 & 79.7 & 67.8 & 72.4 \\
cs-\textsc{attn-ctx} & 1000 & 4 & 66.5 & 74.8 & 85.7 & 84.7 & 70.1 & 36.1 & 88.2 & 70.4 & 79.5 & 66.0 & 72.2 \\
cs-\textsc{attn-attn} & 4000 & 4 & 64.9 & 72.7 & 84.3 & 85.1 & 70.1 & 33.5 & 88.8 & 69.7 & 78.0 & 65.2 & 71.2 \\
cs-\textsc{attn-attn} & 1000 & 4 & 64.0 & 72.6 & 84.6 & 84.2 & 67.9 & 33.2 & 89.0 & 70.1 & 78.0 & 64.6 & 70.8 \\
cs-\textsc{attn-attn} & 1000 & 8 & 62.9 & 71.7 & 83.6 & 84.2 & 67.0 & 34.2 & 86.2 & 69.8 & 76.6 & 63.2 & 70.0 \\
de-\textsc{maxpool-ctx} & 600 & --- & 60.0 & 69.2 & 77.0 & 73.1 & 61.4 & 32.4 & 80.2 & 70.7 & 78.8 & 68.0 & 67.1 \\
de-\textsc{attn-ctx} & 1200 & 12 & 61.1 & 70.0 & 77.3 & 71.7 & 63.5 & 32.4 & 78.4 & 69.8 & 77.4 & 65.0 & 66.7 \\
de-\textsc{attn-ctx} & 600 & 8 & 60.5 & 68.5 & 77.0 & 72.1 & 62.0 & 31.1 & 77.0 & 70.1 & 75.7 & 64.0 & 65.8 \\
de-\textsc{avgpool-ctx} & 600 & --- & 59.5 & 67.5 & 75.6 & 72.5 & 64.1 & 29.3 & 74.6 & 70.8 & 77.5 & 65.2 & 65.6 \\
de-\textsc{attn-ctx} & 600 & 12 & 59.7 & 68.4 & 77.0 & 71.2 & 61.2 & 30.9 & 78.0 & 71.1 & 76.0 & 61.9 & 65.5 \\
de-\textsc{final} & 600 & --- & 59.9 & 65.9 & 76.2 & 72.7 & 61.5 & 31.4 & 73.0 & 70.7 & 77.0 & 64.7 & 65.3 \\
de-\textsc{attn-ctx} & 600 & 3 & 60.3 & 67.0 & 75.4 & 72.7 & 60.6 & 30.4 & 77.0 & 69.9 & 76.0 & 63.3 & 65.3 \\
de-\textsc{attn-attn} & 600 & 1 & 60.0 & 66.5 & 72.8 & 72.2 & 61.7 & 29.5 & 74.2 & 70.5 & 76.9 & 63.8 & 64.8 \\
de-\textsc{attn-attn} & 600 & 3 & 60.7 & 67.5 & 74.1 & 71.8 & 60.6 & 30.1 & 75.0 & 69.5 & 74.7 & 61.5 & 64.5 \\
de-\textsc{final-ctx} & 600 & --- & 58.9 & 66.2 & 73.1 & 71.9 & 61.0 & 29.2 & 75.8 & 70.3 & 76.2 & 62.6 & 64.5 \\
de-\textsc{attn-attn} & 1200 & 6 & 58.7 & 65.9 & 75.4 & 72.3 & 61.0 & 29.7 & 78.4 & 70.1 & 72.3 & 59.6 & 64.3 \\
de-\textsc{trf-attn-attn} & 600 & 3 & 58.8 & 64.9 & 76.2 & 71.7 & 60.3 & 30.4 & 72.0 & 71.2 & 72.5 & 61.4 & 63.9 \\
de-\textsc{attn-attn} & 1200 & 12 & 58.6 & 66.9 & 74.1 & 70.7 & 60.8 & 29.5 & 75.8 & 67.1 & 72.5 & 58.2 & 63.4 \\
de-\textsc{attn-attn} & 2400 & 12 & 57.4 & 66.0 & 74.0 & 70.9 & 58.5 & 27.7 & 76.0 & 67.7 & 73.9 & 59.8 & 63.2 \\
de-\textsc{trf-attn-attn} & 2400 & 12 & 56.9 & 65.3 & 74.4 & 71.2 & 61.2 & 30.5 & 74.0 & 66.1 & 71.2 & 59.0 & 63.0 \\
de-\textsc{attn-attn} & 600 & 6 & 57.4 & 64.8 & 72.4 & 71.8 & 59.5 & 27.2 & 76.0 & 68.6 & 70.9 & 57.5 & 62.6 \\
de-\textsc{attn-attn} & 600 & 8 & 57.5 & 64.5 & 71.7 & 71.8 & 58.8 & 28.1 & 77.4 & 67.0 & 68.6 & 55.6 & 62.1 \\
de-\textsc{trf-attn-attn} & 600 & 6 & 57.8 & 64.6 & 72.0 & 70.8 & 59.3 & 29.2 & 65.6 & 69.1 & 71.0 & 59.5 & 61.9 \\
de-\textsc{attn-attn} & 600 & 12 & 56.0 & 65.6 & 73.1 & 70.5 & 57.6 & 28.6 & 74.0 & 64.1 & 70.5 & 55.2 & 61.5 \\
de-\textsc{trf-attn-attn} & 1200 & 12 & 56.6 & 64.9 & 71.4 & 71.0 & 56.7 & 29.6 & 66.2 & 67.9 & 68.8 & 58.2 & 61.1 \\
de-\textsc{attn-ctx} & 600 & 6 & 58.4 & 63.9 & 72.9 & 70.6 & 57.4 & 29.6 & 58.6 & 66.5 & 68.7 & 62.9 & 61.0 \\
}%
\def\SentEvalB{%
InferSent & 4096 & --- & .88/.83 & .76/.75 & .59/.60 & .59/.59 & .70/.67 & .71/.72 & .71/.73 & .70 \\
cs-\textsc{maxpool} & 1000 & --- & .81/.75 & .72/.71 & .52/.53 & .47/.47 & .54/.53 & .61/.61 & .58/.58 & .60 \\
cs-\textsc{final} & 1000 & --- & .80/.74 & .74/.75 & .54/.56 & .42/.43 & .55/.53 & .60/.59 & .55/.56 & .60 \\
cs-\textsc{final-ctx} & 1000 & --- & .82/.76 & .74/.74 & .51/.53 & .44/.44 & .52/.50 & .62/.61 & .57/.58 & .60 \\
GloVe-BOW & 300 & --- & .80/.72 & .64/.62 & .52/.53 & .50/.51 & .55/.56 & .56/.59 & .51/.58 & .59 \\
cs-\textsc{attn-attn} & 1000 & 1 & .81/.76 & .73/.73 & .46/.49 & .32/.33 & .45/.44 & .53/.52 & .47/.48 & .54 \\
de-\textsc{attn-ctx} & 1200 & 12 & .76/.70 & .52/.51 & .46/.49 & .31/.31 & .50/.50 & .58/.57 & .51/.52 & .52 \\
de-\textsc{attn-ctx} & 600 & 8 & .75/.68 & .52/.50 & .47/.49 & .30/.31 & .52/.52 & .56/.56 & .48/.49 & .51 \\
de-\textsc{attn-attn} & 600 & 1 & .74/.67 & .56/.55 & .46/.48 & .30/.31 & .48/.48 & .53/.53 & .46/.47 & .50 \\
de-\textsc{attn-ctx} & 600 & 3 & .72/.65 & .53/.52 & .45/.48 & .34/.35 & .48/.48 & .55/.54 & .46/.46 & .50 \\
de-\textsc{attn-ctx} & 600 & 12 & .75/.68 & .51/.49 & .46/.47 & .28/.29 & .51/.50 & .54/.54 & .48/.48 & .50 \\
cs-\textsc{avgpool} & 1000 & --- & .78/.72 & .70/.70 & .47/.49 & .29/.30 & .38/.39 & .44/.44 & .41/.43 & .50 \\
de-\textsc{maxpool-ctx} & 600 & --- & .77/.71 & .61/.60 & .46/.48 & .26/.28 & .46/.46 & .51/.52 & .40/.42 & .50 \\
de-\textsc{trf-attn-attn} & 600 & 3 & .70/.63 & .53/.52 & .47/.48 & .31/.31 & .47/.47 & .52/.51 & .47/.47 & .49 \\
de-\textsc{avgpool-ctx} & 600 & --- & .76/.69 & .59/.58 & .44/.46 & .25/.27 & .45/.45 & .50/.49 & .41/.42 & .48 \\
de-\textsc{final-ctx} & 600 & --- & .73/.66 & .57/.55 & .44/.47 & .25/.27 & .43/.43 & .52/.51 & .44/.44 & .48 \\
de-\textsc{final} & 600 & --- & .73/.66 & .62/.60 & .41/.44 & .22/.24 & .43/.43 & .47/.47 & .44/.44 & .47 \\
de-\textsc{attn-attn} & 600 & 3 & .67/.62 & .50/.49 & .44/.47 & .27/.28 & .43/.44 & .50/.49 & .45/.45 & .47 \\
de-\textsc{trf-attn-attn} & 2400 & 12 & .66/.59 & .50/.49 & .42/.42 & .28/.28 & .46/.45 & .51/.51 & .44/.45 & .46 \\
de-\textsc{trf-attn-attn} & 1200 & 12 & .61/.58 & .51/.50 & .44/.46 & .26/.28 & .43/.43 & .50/.50 & .47/.47 & .46 \\
de-\textsc{trf-attn-attn} & 600 & 6 & .66/.59 & .51/.49 & .44/.45 & .27/.28 & .43/.43 & .50/.51 & .39/.41 & .45 \\
cs-\textsc{attn-ctx} & 1000 & 4 & .74/.70 & .64/.64 & .35/.38 & .26/.27 & .31/.31 & .44/.44 & .39/.40 & .45 \\
de-\textsc{attn-attn} & 1200 & 12 & .63/.58 & .40/.39 & .40/.43 & .28/.29 & .40/.41 & .50/.49 & .42/.41 & .43 \\
de-\textsc{attn-ctx} & 600 & 6 & .60/.57 & .47/.47 & .37/.38 & .23/.26 & .42/.43 & .47/.48 & .42/.44 & .43 \\
de-\textsc{attn-attn} & 2400 & 12 & .58/.59 & .40/.39 & .41/.44 & .22/.25 & .39/.39 & .47/.47 & .39/.38 & .41 \\
de-\textsc{attn-attn} & 1200 & 6 & .66/.60 & .39/.39 & .39/.42 & .21/.23 & .37/.37 & .46/.45 & .40/.39 & .41 \\
de-\textsc{attn-attn} & 600 & 12 & .59/.55 & .40/.39 & .39/.43 & .24/.25 & .37/.37 & .46/.46 & .39/.38 & .40 \\
de-\textsc{attn-attn} & 600 & 6 & .61/.56 & .39/.38 & .40/.43 & .22/.23 & .36/.36 & .45/.45 & .38/.37 & .40 \\
de-\textsc{attn-attn} & 600 & 8 & .57/.52 & .37/.36 & .38/.41 & .24/.25 & .35/.36 & .46/.44 & .38/.36 & .39 \\
cs-\textsc{attn-attn} & 1000 & 4 & .70/.66 & .57/.56 & .29/.32 & .22/.21 & .25/.25 & .35/.35 & .34/.34 & .39 \\
cs-\textsc{attn-attn} & 4000 & 4 & .72/.67 & .57/.56 & .29/.32 & .22/.22 & .24/.24 & .36/.35 & .32/.32 & .39 \\
cs-\textsc{attn-attn} & 1000 & 8 & .70/.65 & .54/.52 & .28/.31 & .20/.20 & .22/.22 & .31/.32 & .32/.33 & .36 \\
}%
\def\SentEvalSummary{%
InferSent & 4096 & --- & 83.7 & 86.4 & 81.7 & .70 \\
GloVe-BOW & 300 & --- & 66.0 & 78.2 & 75.8 & .59 \\
cs-\textsc{final-ctx} & 1000 & --- & 70.2 & 82.1 & 74.4 & .60 \\
cs-\textsc{attn-attn} & 1000 & 1 & 69.3 & 80.8 & 73.4 & .54 \\
cs-\textsc{final} & 1000 & --- & 69.2 & 81.1 & 73.2 & .60 \\
cs-\textsc{maxpool} & 1000 & --- & 68.5 & 81.7 & 73.0 & .60 \\
cs-\textsc{avgpool} & 1000 & --- & 67.8 & 79.7 & 72.4 & .50 \\
cs-\textsc{attn-ctx} & 1000 & 4 & 66.0 & 79.5 & 72.2 & .45 \\
cs-\textsc{attn-attn} & 4000 & 4 & 65.2 & 78.0 & 71.2 & .39 \\
cs-\textsc{attn-attn} & 1000 & 4 & 64.6 & 78.0 & 70.8 & .39 \\
cs-\textsc{attn-attn} & 1000 & 8 & 63.2 & 76.6 & 70.0 & .36 \\
de-\textsc{maxpool-ctx} & 600 & --- & 68.0 & 78.8 & 67.1 & .50 \\
de-\textsc{attn-ctx} & 1200 & 12 & 65.0 & 77.4 & 66.7 & .52 \\
de-\textsc{attn-ctx} & 600 & 8 & 64.0 & 75.7 & 65.8 & .51 \\
de-\textsc{avgpool-ctx} & 600 & --- & 65.2 & 77.5 & 65.6 & .48 \\
de-\textsc{attn-ctx} & 600 & 12 & 61.9 & 76.0 & 65.5 & .50 \\
de-\textsc{final} & 600 & --- & 64.7 & 77.0 & 65.3 & .47 \\
de-\textsc{attn-ctx} & 600 & 3 & 63.3 & 76.0 & 65.3 & .50 \\
de-\textsc{attn-attn} & 600 & 1 & 63.8 & 76.9 & 64.8 & .50 \\
de-\textsc{attn-attn} & 600 & 3 & 61.5 & 74.7 & 64.5 & .47 \\
de-\textsc{final-ctx} & 600 & --- & 62.6 & 76.2 & 64.5 & .48 \\
de-\textsc{attn-attn} & 1200 & 6 & 59.6 & 72.3 & 64.3 & .41 \\
de-\textsc{trf-attn-attn} & 600 & 3 & 61.4 & 72.5 & 63.9 & .49 \\
de-\textsc{attn-attn} & 1200 & 12 & 58.2 & 72.5 & 63.4 & .43 \\
de-\textsc{attn-attn} & 2400 & 12 & 59.8 & 73.9 & 63.2 & .41 \\
de-\textsc{trf-attn-attn} & 2400 & 12 & 59.0 & 71.2 & 63.0 & .46 \\
de-\textsc{attn-attn} & 600 & 6 & 57.5 & 70.9 & 62.6 & .40 \\
de-\textsc{attn-attn} & 600 & 8 & 55.6 & 68.6 & 62.1 & .39 \\
de-\textsc{trf-attn-attn} & 600 & 6 & 59.5 & 71.0 & 61.9 & .45 \\
de-\textsc{attn-attn} & 600 & 12 & 55.2 & 70.5 & 61.5 & .40 \\
de-\textsc{trf-attn-attn} & 1200 & 12 & 58.2 & 68.8 & 61.1 & .46 \\
de-\textsc{attn-ctx} & 600 & 6 & 62.9 & 68.7 & 61.0 & .43 \\
}%
\def\ParaEval{%
InferSent & 4096 & --- & 99.99 & 100.00&100.00 & 0.579 & 31.58 & 25.28&26.21 & 0.367 & 48.0 \\
GloVe-BOW & 300 & --- & 99.94 & 100.00&100.00 & 0.654 & 34.28 & 20.29&19.72 & 0.352 & 46.9 \\
cs-\textsc{final-ctx} & 1000 & --- & 99.92 & 100.00&100.00 & 0.406 & 23.20 & 15.74&16.07 & 0.346 & 44.5 \\
cs-\textsc{maxpool} & 1000 & --- & 99.86 & 100.00&100.00 & 0.447 & 21.76 & 15.01&16.34 & 0.348 & 44.2 \\
de-\textsc{attn-ctx} & 600 & 8 & 98.11 & 99.86&99.90 & 0.348 & 21.64 & 15.40&17.32 & 0.349 & 44.1 \\
cs-\textsc{final} & 1000 & --- & 99.91 & 100.00&100.00 & 0.439 & 22.40 & 14.31&14.63 & 0.340 & 44.0 \\
de-\textsc{attn-ctx} & 1200 & 12 & 98.88 & 99.85&99.91 & 0.347 & 20.06 & 14.92&16.68 & 0.348 & 43.9 \\
de-\textsc{maxpool-ctx} & 600 & --- & 98.42 & 99.89&99.90 & 0.343 & 21.54 & 14.65&15.62 & 0.341 & 43.8 \\
de-\textsc{attn-ctx} & 600 & 3 & 97.81 & 99.77&99.87 & 0.328 & 19.74 & 15.28&16.43 & 0.343 & 43.7 \\
de-\textsc{attn-ctx} & 600 & 12 & 97.79 & 99.84&99.89 & 0.360 & 20.22 & 14.54&16.10 & 0.344 & 43.6 \\
de-\textsc{attn-ctx} & 600 & 6 & 98.11 & 99.79&99.86 & 0.358 & 20.44 & 14.48&15.57 & 0.342 & 43.6 \\
de-\textsc{attn-attn} & 600 & 1 & 97.70 & 99.71&99.73 & 0.352 & 19.74 & 14.95&16.26 & 0.340 & 43.6 \\
de-\textsc{avgpool-ctx} & 600 & --- & 97.72 & 99.59&99.60 & 0.312 & 20.04 & 13.49&14.27 & 0.337 & 43.2 \\
cs-\textsc{attn-attn} & 1000 & 1 & 99.88 & 99.91&99.91 & 0.347 & 21.54 & 11.15&11.50 & 0.331 & 43.1 \\
de-\textsc{attn-attn} & 600 & 3 & 97.42 & 99.64&99.75 & 0.314 & 17.36 & 13.35&14.35 & 0.333 & 42.8 \\
de-\textsc{final} & 600 & --- & 97.01 & 99.14&99.30 & 0.305 & 19.88 & 11.41&12.40 & 0.328 & 42.5 \\
de-\textsc{final-ctx} & 600 & --- & 96.65 & 99.66&99.70 & 0.323 & 17.22 & 12.06&12.84 & 0.333 & 42.3 \\
de-\textsc{trf-attn-attn} & 600 & 3 & 95.79 & 99.61&99.64 & 0.315 & 15.76 & 13.20&14.04 & 0.340 & 42.3 \\
cs-\textsc{avgpool} & 1000 & --- & 99.80 & 99.99&99.99 & 0.387 & 17.90 & 8.36&8.61 & 0.311 & 41.9 \\
de-\textsc{attn-attn} & 1200 & 12 & 97.15 & 99.47&99.65 & 0.283 & 12.18 & 11.09&11.97 & 0.330 & 41.5 \\
de-\textsc{attn-attn} & 1200 & 6 & 98.05 & 99.74&99.80 & 0.289 & 11.90 & 9.84&10.69 & 0.327 & 41.3 \\
de-\textsc{attn-attn} & 2400 & 12 & 98.69 & 99.65&99.77 & 0.287 & 10.26 & 9.96&10.94 & 0.326 & 41.2 \\
cs-\textsc{attn-ctx} & 1000 & 4 & 99.75 & 99.72&99.74 & 0.287 & 14.60 & 7.52&7.54 & 0.318 & 41.2 \\
de-\textsc{attn-attn} & 600 & 6 & 96.03 & 99.62&99.71 & 0.287 & 12.22 & 9.92&10.59 & 0.323 & 41.1 \\
de-\textsc{trf-attn-attn} & 2400 & 12 & 95.82 & 99.05&99.03 & 0.307 & 5.66 & 13.85&14.53 & 0.339 & 41.1 \\
de-\textsc{attn-attn} & 600 & 8 & 95.32 & 99.59&99.73 & 0.275 & 10.22 & 9.56&10.58 & 0.325 & 40.7 \\
de-\textsc{attn-attn} & 600 & 12 & 95.16 & 99.52&99.64 & 0.278 & 9.62 & 9.59&10.47 & 0.323 & 40.6 \\
de-\textsc{trf-attn-attn} & 600 & 6 & 90.24 & 98.39&98.44 & 0.313 & 9.06 & 12.98&13.64 & 0.332 & 40.4 \\
de-\textsc{trf-attn-attn} & 1200 & 12 & 90.71 & 98.21&98.22 & 0.301 & 7.06 & 13.10&13.70 & 0.333 & 40.2 \\
cs-\textsc{attn-attn} & 4000 & 4 & 99.54 & 98.89&98.98 & 0.252 & 11.52 & 5.54&5.51 & 0.303 & 40.1 \\
cs-\textsc{attn-attn} & 1000 & 4 & 99.26 & 98.90&98.93 & 0.253 & 10.84 & 5.16&5.20 & 0.299 & 39.9 \\
cs-\textsc{attn-attn} & 1000 & 8 & 99.41 & 98.17&98.09 & 0.243 & 10.24 & 4.51&4.64 & 0.287 & 39.4 \\
}%
\def\ParaEvalSummary{%
InferSent & 4096 & --- & 99.99 & 100.00 & 0.579 & 31.58 & 26.21 & 0.367 \\
GloVe-BOW & 300 & --- & 99.94 & 100.00 & 0.654 & 34.28 & 19.72 & 0.352 \\
cs-\textsc{final-ctx} & 1000 & --- & 99.92 & 100.00 & 0.406 & 23.20 & 16.07 & 0.346 \\
cs-\textsc{maxpool} & 1000 & --- & 99.86 & 100.00 & 0.447 & 21.76 & 16.34 & 0.348 \\
de-\textsc{attn-ctx} & 600 & 8 & 98.11 & 99.90 & 0.348 & 21.64 & 17.32 & 0.349 \\
cs-\textsc{final} & 1000 & --- & 99.91 & 100.00 & 0.439 & 22.40 & 14.63 & 0.340 \\
de-\textsc{attn-ctx} & 1200 & 12 & 98.88 & 99.91 & 0.347 & 20.06 & 16.68 & 0.348 \\
de-\textsc{maxpool-ctx} & 600 & --- & 98.42 & 99.90 & 0.343 & 21.54 & 15.62 & 0.341 \\
de-\textsc{attn-ctx} & 600 & 3 & 97.81 & 99.87 & 0.328 & 19.74 & 16.43 & 0.343 \\
de-\textsc{attn-ctx} & 600 & 12 & 97.79 & 99.89 & 0.360 & 20.22 & 16.10 & 0.344 \\
de-\textsc{attn-ctx} & 600 & 6 & 98.11 & 99.86 & 0.358 & 20.44 & 15.57 & 0.342 \\
de-\textsc{attn-attn} & 600 & 1 & 97.70 & 99.73 & 0.352 & 19.74 & 16.26 & 0.340 \\
de-\textsc{avgpool-ctx} & 600 & --- & 97.72 & 99.60 & 0.312 & 20.04 & 14.27 & 0.337 \\
cs-\textsc{attn-attn} & 1000 & 1 & 99.88 & 99.91 & 0.347 & 21.54 & 11.50 & 0.331 \\
de-\textsc{attn-attn} & 600 & 3 & 97.42 & 99.75 & 0.314 & 17.36 & 14.35 & 0.333 \\
de-\textsc{final} & 600 & --- & 97.01 & 99.30 & 0.305 & 19.88 & 12.40 & 0.328 \\
de-\textsc{final-ctx} & 600 & --- & 96.65 & 99.70 & 0.323 & 17.22 & 12.84 & 0.333 \\
de-\textsc{trf-attn-attn} & 600 & 3 & 95.79 & 99.64 & 0.315 & 15.76 & 14.04 & 0.340 \\
cs-\textsc{avgpool} & 1000 & --- & 99.80 & 99.99 & 0.387 & 17.90 & 8.61 & 0.311 \\
de-\textsc{attn-attn} & 1200 & 12 & 97.15 & 99.65 & 0.283 & 12.18 & 11.97 & 0.330 \\
de-\textsc{attn-attn} & 1200 & 6 & 98.05 & 99.80 & 0.289 & 11.90 & 10.69 & 0.327 \\
de-\textsc{attn-attn} & 2400 & 12 & 98.69 & 99.77 & 0.287 & 10.26 & 10.94 & 0.326 \\
cs-\textsc{attn-ctx} & 1000 & 4 & 99.75 & 99.74 & 0.287 & 14.60 & 7.54 & 0.318 \\
de-\textsc{attn-attn} & 600 & 6 & 96.03 & 99.71 & 0.287 & 12.22 & 10.59 & 0.323 \\
de-\textsc{trf-attn-attn} & 2400 & 12 & 95.82 & 99.03 & 0.307 & 5.66 & 14.53 & 0.339 \\
de-\textsc{attn-attn} & 600 & 8 & 95.32 & 99.73 & 0.275 & 10.22 & 10.58 & 0.325 \\
de-\textsc{attn-attn} & 600 & 12 & 95.16 & 99.64 & 0.278 & 9.62 & 10.47 & 0.323 \\
de-\textsc{trf-attn-attn} & 600 & 6 & 90.24 & 98.44 & 0.313 & 9.06 & 13.64 & 0.332 \\
de-\textsc{trf-attn-attn} & 1200 & 12 & 90.71 & 98.22 & 0.301 & 7.06 & 13.70 & 0.333 \\
cs-\textsc{attn-attn} & 4000 & 4 & 99.54 & 98.98 & 0.252 & 11.52 & 5.51 & 0.303 \\
cs-\textsc{attn-attn} & 1000 & 4 & 99.26 & 98.93 & 0.253 & 10.84 & 5.20 & 0.299 \\
cs-\textsc{attn-attn} & 1000 & 8 & 99.41 & 98.09 & 0.243 & 10.24 & 4.64 & 0.287 \\
}%
\def\SentParaEvalSummary{%
InferSent & 4096 & --- & 83.7 & 86.4 & 81.7 & .70 & 99.99 & 100.00 & 0.579 & 31.58 & 26.21 & 0.367 \\
GloVe-BOW & 300 & --- & 66.0 & 78.2 & 75.8 & .59 & 99.94 & 100.00 & 0.654 & 34.28 & 19.72 & 0.352 \\
cs-\textsc{final-ctx} & 1000 & --- & 70.2 & 82.1 & 74.4 & .60 & 99.92 & 100.00 & 0.406 & 23.20 & 16.07 & 0.346 \\
cs-\textsc{attn-attn} & 1000 & 1 & 69.3 & 80.8 & 73.4 & .54 & 99.88 & 99.91 & 0.347 & 21.54 & 11.50 & 0.331 \\
cs-\textsc{final} & 1000 & --- & 69.2 & 81.1 & 73.2 & .60 & 99.91 & 100.00 & 0.439 & 22.40 & 14.63 & 0.340 \\
cs-\textsc{maxpool} & 1000 & --- & 68.5 & 81.7 & 73.0 & .60 & 99.86 & 100.00 & 0.447 & 21.76 & 16.34 & 0.348 \\
cs-\textsc{avgpool} & 1000 & --- & 67.8 & 79.7 & 72.4 & .50 & 99.80 & 99.99 & 0.387 & 17.90 & 8.61 & 0.311 \\
cs-\textsc{attn-ctx} & 1000 & 4 & 66.0 & 79.5 & 72.2 & .45 & 99.75 & 99.74 & 0.287 & 14.60 & 7.54 & 0.318 \\
cs-\textsc{attn-attn} & 4000 & 4 & 65.2 & 78.0 & 71.2 & .39 & 99.54 & 98.98 & 0.252 & 11.52 & 5.51 & 0.303 \\
cs-\textsc{attn-attn} & 1000 & 4 & 64.6 & 78.0 & 70.8 & .39 & 99.26 & 98.93 & 0.253 & 10.84 & 5.20 & 0.299 \\
cs-\textsc{attn-attn} & 1000 & 8 & 63.2 & 76.6 & 70.0 & .36 & 99.41 & 98.09 & 0.243 & 10.24 & 4.64 & 0.287 \\
de-\textsc{maxpool-ctx} & 600 & --- & 68.0 & 78.8 & 67.1 & .50 & 98.42 & 99.90 & 0.343 & 21.54 & 15.62 & 0.341 \\
de-\textsc{attn-ctx} & 1200 & 12 & 65.0 & 77.4 & 66.7 & .52 & 98.88 & 99.91 & 0.347 & 20.06 & 16.68 & 0.348 \\
de-\textsc{attn-ctx} & 600 & 8 & 64.0 & 75.7 & 65.8 & .51 & 98.11 & 99.90 & 0.348 & 21.64 & 17.32 & 0.349 \\
de-\textsc{avgpool-ctx} & 600 & --- & 65.2 & 77.5 & 65.6 & .48 & 97.72 & 99.60 & 0.312 & 20.04 & 14.27 & 0.337 \\
de-\textsc{attn-ctx} & 600 & 12 & 61.9 & 76.0 & 65.5 & .50 & 97.79 & 99.89 & 0.360 & 20.22 & 16.10 & 0.344 \\
de-\textsc{final} & 600 & --- & 64.7 & 77.0 & 65.3 & .47 & 97.01 & 99.30 & 0.305 & 19.88 & 12.40 & 0.328 \\
de-\textsc{attn-ctx} & 600 & 3 & 63.3 & 76.0 & 65.3 & .50 & 97.81 & 99.87 & 0.328 & 19.74 & 16.43 & 0.343 \\
de-\textsc{attn-attn} & 600 & 1 & 63.8 & 76.9 & 64.8 & .50 & 97.70 & 99.73 & 0.352 & 19.74 & 16.26 & 0.340 \\
de-\textsc{attn-attn} & 600 & 3 & 61.5 & 74.7 & 64.5 & .47 & 97.42 & 99.75 & 0.314 & 17.36 & 14.35 & 0.333 \\
de-\textsc{final-ctx} & 600 & --- & 62.6 & 76.2 & 64.5 & .48 & 96.65 & 99.70 & 0.323 & 17.22 & 12.84 & 0.333 \\
de-\textsc{attn-attn} & 1200 & 6 & 59.6 & 72.3 & 64.3 & .41 & 98.05 & 99.80 & 0.289 & 11.90 & 10.69 & 0.327 \\
de-\textsc{trf-attn-attn} & 600 & 3 & 61.4 & 72.5 & 63.9 & .49 & 95.79 & 99.64 & 0.315 & 15.76 & 14.04 & 0.340 \\
de-\textsc{attn-attn} & 1200 & 12 & 58.2 & 72.5 & 63.4 & .43 & 97.15 & 99.65 & 0.283 & 12.18 & 11.97 & 0.330 \\
de-\textsc{attn-attn} & 2400 & 12 & 59.8 & 73.9 & 63.2 & .41 & 98.69 & 99.77 & 0.287 & 10.26 & 10.94 & 0.326 \\
de-\textsc{trf-attn-attn} & 2400 & 12 & 59.0 & 71.2 & 63.0 & .46 & 95.82 & 99.03 & 0.307 & 5.66 & 14.53 & 0.339 \\
de-\textsc{attn-attn} & 600 & 6 & 57.5 & 70.9 & 62.6 & .40 & 96.03 & 99.71 & 0.287 & 12.22 & 10.59 & 0.323 \\
de-\textsc{attn-attn} & 600 & 8 & 55.6 & 68.6 & 62.1 & .39 & 95.32 & 99.73 & 0.275 & 10.22 & 10.58 & 0.325 \\
de-\textsc{trf-attn-attn} & 600 & 6 & 59.5 & 71.0 & 61.9 & .45 & 90.24 & 98.44 & 0.313 & 9.06 & 13.64 & 0.332 \\
de-\textsc{attn-attn} & 600 & 12 & 55.2 & 70.5 & 61.5 & .40 & 95.16 & 99.64 & 0.278 & 9.62 & 10.47 & 0.323 \\
de-\textsc{trf-attn-attn} & 1200 & 12 & 58.2 & 68.8 & 61.1 & .46 & 90.71 & 98.22 & 0.301 & 7.06 & 13.70 & 0.333 \\
de-\textsc{attn-ctx} & 600 & 6 & 62.9 & 68.7 & 61.0 & .43 & 98.11 & 99.86 & 0.358 & 20.44 & 15.57 & 0.342 \\
}%
\def\csReprCorrelationMean{0.78\pm0.32}
\def\csReprTRECCorrelationMean{-0.16\pm0.16}
\def\csReprBLEUCorrelationMean{-0.57\pm0.31}
\def\csReprHeadsCorrelationMean{-0.81\pm0.12}
\def\deReprHeadsCorrelationMean{-0.18\pm0.19}
\def\csHeadsBLEUCorrelation{0.87}
\def\deHeadsBLEUCorrelation{0.31}
 % load generated tables

\section{Introduction}

Deep learning has brought the possibility of
automatically learning continuous representations of sentences. On the one hand,
such
representations can be geared towards particular tasks such as classifying the
sentence in various aspects (e.g.\ sentiment, register, question type) or relating
the sentence to other sentences (e.g.\ semantic similarity, paraphrasing,
entailment). On the other hand, we can aim at \equo{universal} sentence
representations, that is representations performing reasonably well in a range of such tasks.

Regardless the evaluation criterion, the representations can be learned either in an
unsupervised way (from simple, unannotated texts) or supervised, relying on
manually constructed training sets of sentences equipped with annotations of the
appropriate type. A different approach is to obtain sentence representations from training neural machine translation models
%, which has been explored by \citet{Hill2016LearningDR}.
\parcite{Hill2016LearningDR}.

Since \perscite{Hill2016LearningDR}, NMT has seen substantial advances in
translation quality
and it is thus natural to ask how these improvements affect the learned
representations.

One of the key technological changes was the introduction of \equo{attention}
\parcite{Bahdanau2014NeuralMT}, making it even the very central component in the
network \parcite{Vaswani2017AttentionIA}. Attention allows the NMT system to
dynamically choose which parts of the source are most important when deciding on
the current output token. As a consequence, there is no longer a static vector
representation of the sentence available in the system.

In this paper, we remove this limitation by proposing a novel encoder-decoder
architecture with a structured fixed-size representation of the input that
still allows the decoder to explicitly focus on different parts of the input. In
other words, our NMT system has both the capacity to attend to various parts of
the input and to produce static representations of input sentences.

We train this architecture on English-to-German and English-to-Czech translation
and evaluate the learned representations of English on a wide
range of tasks in order to assess its performance in
learning \equo{universal}
meaning representations.

\begin{table*}[t]
\centering
\small
\begin{tabular}{l|c|cccccc}
\multicolumn{1}{c}{} &
\multicolumn{1}{c}{\clap{\tiny{\citeauthor{Bahdanau2014NeuralMT}}}} &
\tiny{\citeauthor{Sutskever2014SequenceTS}} & \tiny{\citeauthor{Cho2014LearningPR}} & & & & \clap{\tiny{Compound attention}}\\
& \textsc{attn} & \textsc{final} & \textsc{final-ctx} & \textsc{*pool} & \textsc{*pool-ctx} & \textsc{attn-ctx} & \textsc{attn-attn} \\
%\midrule
\hline
Encoder states used & all & final & final & all & all & all & all \\
Combined using \dots & --- & --- & --- & pooling & pooling & inner att. & inner att. \\
Sent. emb. available & \xmark & \cmark & \cmark & \cmark & \cmark & \cmark & \cmark \\
\hline
Dec.\ attends to enc.\ states & \cmark & \xmark & \xmark & \xmark & \xmark & \xmark & \xmark \\
Dec.\ attends to sent.\ emb. & \xmark & \xmark & \xmark & \xmark & \xmark & \xmark & \cmark \\
Sent.\ emb.\ used in \dots & --- & init & init+ctx & init & init+ctx & init+ctx & input for att.
\end{tabular}
\caption{Different RNN-based architectures and their properties. Legend:
\equo{pooling} -- vectors combined by mean or max (\textsc{avgpool},
\textsc{maxpool}); \equo{sent. emb.} -- sentence embedding, i.e.\ the
fixed-size sentence representation
computed by the encoder.
\equo{init} -- initial decoder state. \equo{ctx} --
context vector, i.e.\ input for the decoder cell.
\equo{input for att.} -- input for
the decoder attention.}
\label{tab:architectures}
\end{table*}

In \cref{related}, we briefly review recent efforts in obtaining sentence
representations. In \cref{arch}, we introduce a number of variants of our novel
architecture. \cref{eval} describes some standard and our own methods for evaluating
sentence representations. \cref{experiments} then provides experimental results:
translation and representation quality. The relation between the two is discussed in \cref{discussion}.
%We conclude in \cref{conclusion}.

\section{Related Work}
\label{related}

The properties of continuous sentence representations have always been of interest to researchers working on neural machine translation. In the first works on RNN sequence-to-sequence models, \citet{Cho2014LearningPR} and \citet{Sutskever2014SequenceTS} provided visualizations of the phrase and sentence embedding spaces and observed that they reflect semantic and syntactic structure to some extent.

\citet{Hill2016LearningDR} perform a systematic evaluation of sentence representation in different models, including NMT, by applying them to various sentence classification tasks and by relating semantic similarity to closeness in the representation space.

\citet{Shi2016DoesSN} investigate the syntactic properties of
representations learned by NMT systems by predicting sentence- and
word-level syntactic labels (e.g.\ tense, part of speech) and by generating
syntax trees from these representations.

\citet{Schwenk2017LearningJM} aim to learn language-independent sentence
representations using NMT systems with multiple source and target languages.
They do not consider the attention mechanism and evaluate primarily by similarity scores of the learned representations for similar sentences (within or across languages).

\section{Model Architectures}
\label{arch}

%We now describe our proposed NMT model architectures. These architectures differ in
Our proposed model architectures differ in
\begin{enumerate*}[label=(\alph*)]
\item which encoder states are considered in subsequent processing,
\item how they are combined, and
\item how they are used in the decoder.
\end{enumerate*}

\cref{tab:architectures} summarizes all the examined configurations of
RNN-based models.
The first three (\textsc{attn}, \textsc{final}, \textsc{final-ctx}) correspond roughly to the standard
sequence-to-sequence models, \perscite{Bahdanau2014NeuralMT}, \perscite{Sutskever2014SequenceTS} and
\perscite{Cho2014LearningPR}, respectively. The last column (\textsc{attn-attn}) is our
main proposed architecture: compound attention, described here in
\cref{compound-attn}.

In addition to RNN-based models, we  experiment with the Transformer model,
see \cref{transformer-inner-attn}.

\subsection{Compound Attention}
\label{compound-attn}

\begin{figure}[t]
\centering
\includegraphics{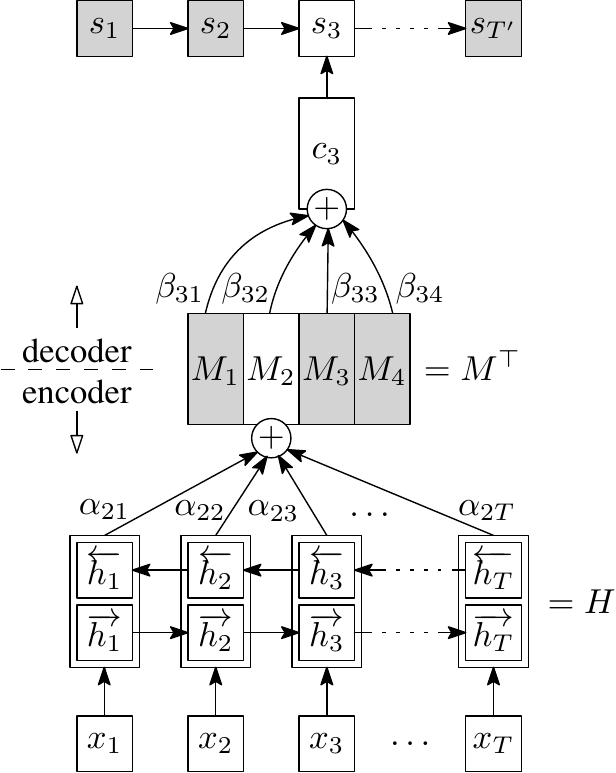}
\caption{An illustration of compound attention with 4 attention heads. The
figure shows the computations that result in the decoder state $s_3$ (in
addition, each state $s_i$ depends on the previous target token $y_{i-1}$). Note
that the matrix $M$ is the same for all positions in the output
sentence and it can thus serve as the source sentence
representation.}
\label{fig:arch}
\end{figure}

Our compound attention model incorporates attention in both the encoder and the decoder. Its architecture is shown in \cref{fig:arch}.

\paragraph{Encoder with inner attention.}
First, we process the input sequence $x_1,x_2,\ldots,x_T$ using a bi-directional recurrent network with gated recurrent units (GRU, \citealp{Cho2014LearningPR}):
\begin{align*}
\overrightarrow{h_t} &= \overrightarrow{\text{GRU}}(x_t, \overrightarrow{h_{t-1}}), \\
\overleftarrow{h_t} &= \overleftarrow{\text{GRU}}(x_t, \overleftarrow{h_{t+1}}), \\
h_t &= [\overrightarrow{h_t}, \overleftarrow{h_t}].
\end{align*}
We denote by $u$ the combined number of units in the two
RNNs, i.e.\ the dimensionality of $h_t$.

Next, our goal is to combine the states $(h_1,h_2,\ldots,h_T)=H$ of the encoder into a vector of fixed dimensionality that represents the entire sentence.
Traditional seq2seq models concatenate the final states of both encoder
RNNs
($\overrightarrow{h_T}$ and $\overleftarrow{h_1}$) to obtain the sentence
representation (denoted as \textsc{final} in \cref{tab:architectures}).
Another option is to combine all encoder states using the average or maximum
over time \citep{Collobert2008AUA,Schwenk2017LearningJM}
(\textsc{avgpool} and \textsc{maxpool} in \cref{tab:architectures} and
following).
 
We adopt an alternative approach, which is to use \emph{inner attention}%
\footnote{Some papers call the same or similar approach \emph{self-attention} or \emph{single-time attention}.}
\citep{Liu2016LearningNL,Li2016DatasetAN} to compute several weighted averages of the encoder states \citep{Lin2017ASS}.
The main motivation for incorporating these multiple \equo{views} of the state
sequence is that it removes the need for the RNN cell to accumulate the
representation of the whole sentence as it processes the input, and therefore it should have more capacity for modeling local dependencies.

Specifically, we fix a number $r$, the number of \emph{attention heads}, and compute an $r\times T$ matrix $A$ of attention weights $\alpha_{jt}$, representing the importance of position $t$ in the input for the $j$\textsuperscript{th} attention head.
We then use this matrix to compute $r$ weighted sums of the encoder states, which become the rows of a new matrix $M$:
\begin{equation}
M = AH. \label{eq:att-weighting-1}
\end{equation}
A vector representation of the source sentence (the \equo{sentence embedding}) can be obtained by flattening the matrix $M$.
In our experiments, we project the encoder states $h_1,h_2,\ldots,h_T$ down to a given dimensionality before applying \cref{eq:att-weighting-1}, so that we can control the size of the representation.

Following \citet{Lin2017ASS}, we compute the attention matrix by feeding the encoder states to a two-layer feed-forward network:
\begin{equation}
A = \softmax(U\tanh(W H^\top)),
\end{equation}
where $W$ and $U$ are weight matrices of dimensions $d\times u$ and $r\times d$, respectively ($d$ is the number of hidden units);
the softmax function is applied along the second dimension, i.e.\ across the encoder states.

\paragraph{Attentive decoder.}
\label{sec:att-decoder}
In vanilla seq2seq models with a fixed-size sentence representation, the decoder is usually conditioned on this representation via the initial RNN state.
We propose to instead leverage the structured sentence embedding by applying attention to its components.
This is no different from the classical attention mechanism used in NMT \citep{Bahdanau2014NeuralMT}, except that it acts on this fixed-size representation instead of the sequence of encoder states.

In the $i$\textsuperscript{th} decoding step, the attention mechanism computes a distribution $\{\beta_{ij}\}_{j=1}^{r}$ over the $r$ components of the structured representation. This is then used to weight these components to obtain the context vector $c_i$, which in turn is used to update the decoder state. Again, we can write this in matrix form as
\begin{equation}
C = BM, \label{eq:att-weighting-2}
\end{equation}
where $B=(\beta_{ij})_{i=1,j=1}^{T',r}$ is the attention matrix and $C=(c_i,c_2,\ldots,c_{T'})$ are the context vectors.

Note that by combining \cref{eq:att-weighting-1,eq:att-weighting-2}, we get
\begin{equation}
C=(BA)H.
\end{equation}
Hence, the composition of the encoder and decoder attentions (the \equo{compound
attention}) defines an implicit alignment between the source and the target sequence. From this viewpoint, our model can be regarded as a restriction of the conventional attention model.

The decoder uses a conditional GRU cell (cGRU\textsubscript{att}; \citealp{Sennrich2017NematusAT}), which consists of two consecutively applied GRU blocks. The first block processes the previous target token $y_{i-1}$, while the second block receives the context vector $c_i$ and predicts the next target token $y_i$.

\begin{table*}[t]
\centering\small
\begin{tabularx}{\textwidth}{llrrlX}
%\multirow{2}{*}[-2pt]{Name} & \multirow{2}{*}[-2pt]{Cl.} & \multicolumn{2}{c}{Data size} & \multirow{2}{*}[-2pt]{Task} & \multirow{2}{*}[-2pt]{Example}\\
%& & \multicolumn{1}{c}{train} & \multicolumn{1}{c}{test} \\
Name & Cl. & Train & Test & Task & Example Input and Label \\
\hline
MR & 2 & 11k & --- & sentiment (movies) & \textit{an idealistic love story that brings out the latent 15-year-old romantic in everyone.} ($+$)\\
CR & 2 & 4k & --- & product review polarity & \textit{no way to contact their customer service.} ($-$)\\
SUBJ & 2 & 10k & --- & subjectivity & \textit{a little weak -- and it isn't that funny.} (subjective)\\
MPQA & 2 & 11k & --- & opinion polarity & \textit{was hoping} ($+$), \textit{breach of the very constitution} ($-$) \\
SST2 & 2 & 68k & 2k & sentiment (movies) & \textit{contains very few laughs and even less surprises} ($-$) \\
SST5 & 5 & 10k & 2k & sentiment (movies) & \textit{it's worth taking the kids to.} (4)\\
TREC & 6 & 5k & 500 & question type & \textit{What was Einstein s IQ?} (NUM)\\
MRPC & 2 & 4k & 2k & semantic equivalence & \textit{Lawtey is not the first faith-based program in Florida's prison system.} / \textit{But Lawtey is the first entire prison to take
that path.} ($-$)\\
SNLI & 3 & 559k & 10k & natural language inference & \textit{Two doctors perform surgery on patient.} / \textit{Two surgeons are having lunch.} (contradiction)\\
SICK-E & 3 & 5k & 5k & natural language inference & \textit{A group of people is near the ocean} / \textit{A crowd of people is near the water} (entailment)\\
\hline
\end{tabularx}
\caption{SentEval classification tasks.
Tasks without a test set use 10-fold cross-validation.}
\label{tab:class-tasks}
\end{table*}

\subsection{Constant Context}

Compared to the \textsc{final} model, the compound attention architecture described in the previous section undoubtedly benefits from the fact that the decoder is presented with information from the encoder (i.e.\ the context vectors $c_i$) in every decoding step.
To investigate this effect, we include baseline models where we replace all context vectors $c_i$ with the entire sentence embedding (indicated
by the suffix
\equo{\textsc{-ctx}} in \cref{tab:architectures}).
Specifically, we provide either the flattened matrix $M$ (for models with inner
attention; \textsc{attn} or \textsc{attn-ctx}), the final state of the encoder (\textsc{final-ctx}), or the result of mean- or
max-pooling (\textsc{*pool-ctx}) as a constant input to the decoder cell.

\subsection{Transformer with Inner Attention}
\label{transformer-inner-attn}

The Transformer \citep{Vaswani2017AttentionIA} is a recently proposed model based entirely on feed-forward layers and attention. It consists of an encoder and a decoder, each with 6 layers, consisting of multi-head attention on the previous layer and a position-wise feed-forward network.

In order to introduce a fixed-size sentence representation into the model, we
modify it by adding inner attention after the last encoder layer. The attention
in the decoder then operates on the components of this representation (i.e.\ the rows
of the matrix $M$).
This variation on the Transformer model corresponds to the \textsc{attn-attn}
column in \Tref{tab:architectures} and is therefore denoted \textsc{trf-attn-attn}.

\section{Representation Evaluation}
\label{eval}

Continuous sentence representations can be evaluated in many ways, see e.g.\ 
\perscite{skip-thought:2015}, \perscite{Conneau2017SupervisedLO} or the %series of
RepEval workshops.\footurl{https://repeval2017.github.io/}

We evaluate our learned representations with classification and similarity tasks
from SentEval (\cref{senteval}) and by examining clusters of sentence
paraphrase representations (\cref{paraphrases}).

\subsection{SentEval}
\label{senteval}

%paraphrase identification (MSRP)
% (Dolan et al., 2004), movie review sentiment
% (MR) (Pang and Lee, 2005), product reviews
% (CR) (Hu and Liu, 2004), subjectivity classification
% (SUBJ) (Pang and Lee, 2004), opinion polarity
% (MPQA) (Wiebe et al., 2005) and question type
% classification (TREC) (Voorhees, 2002). We follow
% the procedure (and code) of Kiros et al. (2015):

\begin{table}
\centering\small
\begin{tabular}{lrrl}
%\multirow{2}{*}[-2pt]{Name} & \multicolumn{2}{c}{Data Size} & \multirow{2}{*}[-2pt]{Method}\\
%& \multicolumn{1}{c}{Train} & \multicolumn{1}{c}{Test} \\
Name & Train & Test & Method\\
\hline
SICK-R & 5k & 5k & regression \\
STSB & 7k & 1k & regression \\
STS12 & --- & 3k & cosine similarity \\
STS13 & --- & 2k & cosine similarity \\
STS14 & --- & 4k & cosine similarity \\
STS15 & --- & 9k & cosine similarity \\
STS16 & --- & 9k & cosine similarity \\
\hline
\end{tabular}
\caption{SentEval semantic relatedness tasks.
%\equo{Train} includes validation data. % zamlcuju, neni zasadni
}
\label{tab:sim-tasks}
\end{table}

We perform evaluation on 10 classification and 7 similarity tasks using the SentEval\footnote{\url{https://github.com/facebookresearch/SentEval/}} \citep{Conneau2017SupervisedLO} evaluation tool. This is a superset of the tasks from \citet{skip-thought:2015}.

\cref{tab:class-tasks} describes the classification tasks (number of classes,
data size, task type and an example) and \cref{tab:sim-tasks} lists the
similarity tasks. The similarity (relatedness) datasets contain pairs of sentences labeled with a real-valued similarity score. A given sentence
representation model is evaluated either by learning to directly
predict this score given the respective sentence embeddings (\equo{regression}), or by computing the
cosine similarity of the embeddings (\equo{similarity}) without the need of any training.
In both cases, Pearson and Spearman correlation of the predictions with the gold
ratings is reported.

See \citet{Dolan2004UnsupervisedCO} for details on MRPC and \citet{Hill2016LearningDR} for the remaining tasks.

\subsection{Paraphrases}
\label{paraphrases}

We also evaluate the representation of paraphrases. We use two paraphrase
sources for this purpose: COCO and HyTER Networks.

COCO (Common Objects in Context; \citealp{LinMBHPRDZ14}) is an object
recognition and image captioning dataset, containing 5 captions for each image.
We extracted the captions from its validation set to form a set of
$5\times5\text{k}=25\text{k}$ sentences grouped by the source image.
The average sentence length is 11 tokens and the vocabulary
size is 9k %token
types.

HyTER Networks \citep{LDC2014T09} are large finite-state
networks
representing a subset of all possible English translations of 102 Arabic and
102 Chinese sentences. The
networks were built by manually  based on reference sentences
in Arabic, Chinese and English. Each network contains up to hundreds of thousands of
possible translations of the given source sentence. We randomly generated 500 translations for each
source sentence, obtaining a corpus of 102k sentences grouped into 204 clusters, each containing 500 paraphrases. The average length of the 102k English sentences is 28 tokens and the vocabulary size is 11k token types.

For every model, we encode each dataset to obtain a set of sentence embeddings with cluster labels. We then compute the following metrics:

%\begin{enumerate}[label=(\roman*)]
%\item
\textbf{Cluster classification accuracy} (denoted \equo{Cl}). We remove 1 point (COCO) or half of the points (HyTER) from each cluster, and fit an LDA classifier on the rest. We then compute the accuracy of the classifier on the removed points.

%\item
\textbf{Nearest-neighbor paraphrase retrieval accuracy} (NN). For each point, we find its nearest neighbor according to cosine or $L_2$ distance, and count how often the neighbor lies in the same cluster as the original point.

%\item
\textbf{Inverse Davies-Bouldin index} (iDB).
The Davies-Bouldin index \parcite{Davies1979ACS} measures cluster
separation. For every pair of clusters, we compute the ratio $R_{ij}$ of their combined
scatter (average $L_2$ distance to the centroid) $S_i+S_j$ and the $L_2$ distance of their
centroids $d_{ij}$, then average the maximum values for all clusters:
\begin{gather}
R_{ij} = \frac{S_i + S_j}{d_{ij}} \\
\text{DB} = \frac{1}{N} \sum_{i=1}^{N} \max_{j\neq i} R_{ij}
\end{gather}
The lower the DB index, the better the separation. To match with the rest of
our metrics, we take its inverse:
%\begin{equation}
$
\text{iDB} = \frac{1}{\text{DB}}
$.
%\end{equation}
%\end{enumerate}

\section{Experimental Results}
\label{experiments}

\def\cs{\textit{cs}}
\def\de{\textit{de}}

We trained English-to-German and English-to-Czech NMT models using Neural
Monkey\footnote{\url{https://github.com/ufal/neuralmonkey}}
\citep{helcl2017neural}. In the following, we distinguish these models using the
code of the target language, i.e.\ \de{} or \cs{}.

The \de{} models were trained on the Multi30K multilingual image
caption dataset \citep{Elliott2016Multi30KME}, extended by
\citet{Helcl2017CUNISF}, who acquired additional parallel data using
back-translation \citep{Sennrich2016ImprovingNM} and perplexity-based selection
\citep{Yasuda2008MethodOS}. This extended dataset contains 410k
sentence pairs, with the average sentence length of $12\pm4$
tokens in English. We train
each model for 20 epochs with
the batch size of 32. We truecased the training data as well as all
data we evaluate on. For German, we employed Neural Monkey's reversible
pre-processing scheme, which expands contractions and performs morphological
segmentation of determiners. We used a vocabulary of at most 30k tokens for each
language (no subword units).

The \cs{} models were trained on CzEng 1.7
\parcite{czeng16:2016}.\footurl{http://ufal.mff.cuni.cz/czeng/czeng17} We used
byte-pair encoding (BPE) with a vocabulary of 30k sub-word units, shared for
both languages. For English, the average sentence length is $15\pm19$
BPE tokens
and the original vocabulary size is 1.9M.
We performed 1 training epoch with the batch size of 128 on the
entire training section (57M sentence pairs).

The datasets for both \de{} and \cs{} models come with their respective
development and test sets of sentence pairs, which we use for the evaluation of
translation quality. (We use 1k randomly selected sentence pairs from
CzEng 1.7 dtest as a development set. For evaluation, we use the entire etest.)

We also evaluate the InferSent
model\footnote{\url{https://github.com/facebookresearch/InferSent}}
\citep{Conneau2017SupervisedLO} as pre-trained on the natural language
inference (NLI) task. InferSent has been shown to achieve state-of-the-art results on
the SentEval tasks.
We also include a bag-of-words baseline (GloVe-BOW) obtained by averaging GloVe\footurl{https://nlp.stanford.edu/projects/glove/} word vectors \citep{pennington2014glove}.

\subsection{Translation Quality}
\begin{table}[t]
\centering\small
\begin{tabular}{l@{~~}r@{~~}r|r@{~~}r}
%\toprule
\multirow{2}{*}{Model} & 
\multirow{2}{*}{Size} & 
\multirow{2}{*}{Heads} & \multicolumn{2}{c}{BLEU} \\
& & & \multicolumn{1}{c}{dev} & \multicolumn{1}{c}{test} \\
\hline
\BLEUmmm
%\bottomrule
\end{tabular}
\caption{Translation quality of \de{} models.}
\label{tab:deBLEU}
\end{table}

\begin{table}[t]
\centering\small
\begin{tabular}{l@{~~}r@{~~}r|r@{~~}r|r}
%\toprule
\multirow{2}{*}{Model} & 
\multirow{2}{*}{Size} & 
\multirow{2}{*}{Heads} & \multicolumn{2}{c|}{BLEU} & Manual \\
& & & \multicolumn{1}{c}{dev} & \multicolumn{1}{c|}{test} & \multicolumn{1}{c}{$>$ others} \\
\hline
cs-\textsc{attn} & --- & --- & 22.8 & 22.2 & 89.1 \\
cs-\textsc{attn-attn} & 1000 & 8 & 19.1 & 18.4 & 78.8 \\
cs-\textsc{attn-attn} & 4000 & 4 & 18.4 & 17.9 & --- \\
cs-\textsc{attn-attn} & 1000 & 4 & 17.5 & 17.1 & --- \\
cs-\textsc{attn-ctx} & 1000 & 4 & 16.6 & 16.1 & 58.8 \\
cs-\textsc{final-ctx} & 1000 & --- & 16.1 & 15.5 & --- \\
cs-\textsc{attn-attn} & 1000 & 1 & 15.3 & 14.8 & 49.1 \\
cs-\textsc{final} & 1000 & --- & 11.2 & 10.8 & --- \\
cs-\textsc{avgpool} & 1000 & --- & 11.1 & 10.6 & --- \\
cs-\textsc{maxpool} & 1000 & --- & 5.4 & 5.4 & 3.0 \\
\end{tabular}
\caption{Translation quality of \cs{} models.}
\label{tab:csBLEU}
\end{table}

We estimate translation quality of the various models using single-reference
case-sensitive
BLEU \parcite{papineni:2002} as implemented in Neural Monkey (the reference implementation is \texttt{mteval-v13a.pl} from NIST or Moses).

\cref{tab:deBLEU,tab:csBLEU} provide the results on the two
datasets. The \cs{} dataset is much larger and the training takes much longer.
We were thus able to experiment with only a subset of the possible model
configurations.

The columns \equo{Size} and \equo{Heads} specify the total size of sentence
representation and the number of heads of encoder inner attention.

In both cases, the best performing is the \textsc{attn}
\citeauthor{Bahdanau2014NeuralMT} model, followed by Transformer (\de{} only)
and our \textsc{attn-attn} (compound attention). The non-attentive
\textsc{final} \citeauthor{Cho2014LearningPR} is the worst, except
cs-\textsc{maxpool}.

For 5 selected \cs{} models, we also performed the WMT-style 5-way manual ranking on 200 sentence pairs. The annotations are interpreted as simulated pairwise comparisons. For each model, the final score is the number of times the model was judged better than the other model in the pair. Tied pairs are excluded.
The results, shown in \cref{tab:csBLEU}, confirm the automatic evaluation results.

We also checked the relation between BLEU and the number of heads and
representation size.
While there are many exceptions, the
general tendency is that the larger the representation or the more
heads, the higher the BLEU score.
The Pearson correlation between BLEU and the number of heads is 0.87 for \cs{} and 0.31 for \de{}.

\begin{table*}
\centering
\small
\linespread{.96}\selectfont
\begin{tabular}{l@{~~}r@{~~}r|r@{~}r@{~}r@{~}r||r@{~~}r@{~~}r|r@{~~}r@{~~}r}
Name &  Size & H. & SNLI & SICK-E & AvgAcc & AvgSim & Hy-Cl & Hy-NN & Hy-iDB &  CO-Cl & CO-NN & CO-iDB\\
\hline
\SentParaEvalSummary
\hline
\multicolumn{3}{l|}{LM perplexity (\cs{})} & 190.6 & 299.4  & 1150.2 & 1224.2 & \multicolumn{3}{c|}{668.5}  & \multicolumn{3}{c}{238.5}\\
\multicolumn{3}{l|}{\% OOV (\cs{})} &  0.3 & 0.2  & 2.3 & 2.6 & \multicolumn{3}{c|}{1.2}  & \multicolumn{3}{c}{0.1}\\
\multicolumn{3}{l|}{LM perplexity (\de{})} & 38.8  & 65.0 & 3578.2 & 2010.6 & \multicolumn{3}{c|}{3354.8}  & \multicolumn{3}{c}{86.3}\\
\multicolumn{3}{l|}{\% OOV (\de{})} &  1.5 & 1.7  & 17.8 & 16.2 & \multicolumn{3}{c|}{19.3}  & \multicolumn{3}{c}{1.9}\\
\end{tabular}
\caption{Abridged SentEval and paraphrase evaluation results.
Full results in supplementary material. AvgAcc is the average of all 10 SentEval classification tasks (see \subtref{tab:acc-full}), AvgSim averages all 7 similarity tasks (see \cref{tab:sim-full}). Hy- and CO- stand for HyTER and COCO, respectively. \equo{H.} is the number of attention heads. We give the out-of-vocabulary (OOV) rate and the perplexity  of a 4-gram language model (LM) trained on the English side of the respective parallel corpus and evaluated on all available data for a given task.}
\label{tab:senteval-para}
\end{table*}

\begin{table*}
\centering
{
%\linespread{0.7}\selectfont
\begin{tabular*}{\linewidth}{l@{~~}r@{~~}r|r@{\extracolsep{\stretch{1}}}@{~}r@{~}r@{~}r@{~}r@{~}r@{~}r@{~}r@{~}r@{~}r@{~}r}
\hline
Name &  Size & H. & \small MR & \small CR &  \small \small SUBJ & \small MPQA & \small SST2 & \small SST5 & \small TREC & \small MRPC & \small SICK-E & \small SNLI & \small AvgAcc \\
\hline
\multicolumn{3}{l}{Most frequent baseline} & 50.0 & 63.8 & 50.0 & 68.8 & 49.9 & 23.1 & 18.8 & 66.5 & 56.7 & 34.3 & 48.2 \\
\hline
%InferSent$^\dagger$ & 4096 & --- & 81.1 & 86.3 & 92.4 & 90.2 & 84.6 & --- & 88.2 & 76.2 & 86.3 & \parens{84.5} & --- \\
InferSent & 4096 & --- & \tbf 81.5 & \tbf 86.7 & 92.7 & \tbf 90.6 & \tbf 85.0 & \tbf 45.8 & 88.2 & 76.6 & \tbf 86.4 & \parens{83.7} & \tbf 81.7 \\
\citeauthor{Hill2016LearningDR} en$\rightarrow$fr$^\dagger$  & 2400 & --- & 64.7 & 70.1 & 84.9 & 81.5 & --- & --- & 82.8 & \tbf 96.1 & --- & --- & --- \\
SkipThought-LN$^\dagger$ & --- & --- & 79.4 & 83.1 & \tbf 93.7 & 89.3 & 82.9 & --- & 88.4 & --- & 79.5 & --- & --- \\
GloVe-BOW & 300 & --- & 77.0 & 78.2 & 91.1 & 87.9 & 81.0 & 44.4 & 82.0 & 72.3 & 78.2 & 66.0 & 75.8 \\
cs-\textsc{final-ctx} & 1000 & --- & 68.7 & 77.4 & 88.5 & 85.5 & 73.0 & 38.2 & 88.6 & 71.8 & 82.1 & 70.2 & 74.4 \\
cs-\textsc{attn-attn} & 1000 & 1 & 68.2 & 76.0 & 86.9 & 84.9 & 72.0 & 35.7 & \tbf 89.0 & 70.7 & 80.8 & 69.3 & 73.4 \\
\hline
\end{tabular*}
\begin{tabular*}{\linewidth}{l@{~~}r@{~~}r|r@{\extracolsep{\stretch{1}}}@{~}r@{~}r@{~}r@{~}r@{~}r@{~}r@{~}r@{~}}
\hline
Name &  Size & H. & \small SICK-R & \small STSB & \small STS12 & \small STS13 & \small STS14 & \small STS15 & \small STS16 & \small AvgSim \\
\hline
%InferSent$^\dagger$ & 4096 & --- & 76.2 & 86.3 & \parens{84.5} & --- \\
InferSent & 4096 & --- & {\tbf.88}/{\tbf.83} & {\tbf.76}/{\tbf.75} & {\tbf.59}/{\tbf.60} & {\tbf.59}/{\tbf.59} & {\tbf.70}/{\tbf.67} & {\tbf.71}/{\tbf.72} & {\tbf.71}/{\tbf.73} & \tbf .70 \\
SkipThought-LN$^\dagger$ & --- & --- & \makebox[\widthof{.88/.83}][l]{.85/\,\,---} & --- & --- & --- & .44/.45 & --- & --- & --- \\
GloVe-BOW & 300 & --- & .80/.72 & .64/.62 & .52/.53 & .50/.51 & .55/.56 & .56/.59 & .51/.58 & .59 \\
cs-\textsc{final-ctx} & 1000 & --- & .82/.76 & .74/.74 & .51/.53 & .44/.44 & .52/.50 & .62/.61 & .57/.58 & .60 \\
cs-\textsc{attn-attn} & 1000 & 1 & .81/.76 & .73/.73 & .46/.49 & .32/.33 & .45/.44 & .53/.52 & .47/.48 & .54 \\
\hline
\end{tabular*}
}
\caption{Comparison of state-of-the-art SentEval results with our best models and the Glove-BOW baseline. \equo{H.} is the number of attention heads. Reprinted results are marked with $\dagger$, others are our measurements.
}
\label{tab:senteval-truncated}
\end{table*}

\subsection{SentEval}
Due to the large number of SentEval tasks, we present the results abridged in two different ways: by reporting averages (\cref{tab:senteval-para}) and by showing only the best models in comparison with other methods (\cref{tab:senteval-truncated}).
The full results can be found in the supplementary material.

\cref{tab:senteval-para} provides averages of the classification and similarity results, along with the results of selected tasks (SNLI, SICK-E). As the baseline for classifications tasks, we assign the most frequent class to all test
examples.\footnote{For MR, CR, SUBJ, and MPQA, where there is no distinct
test set, the class is established on the whole collection. For other tasks, the
class is learned from the training set.} 
The \de{} models are
generally worse, most likely due to the higher OOV rate and overall
simplicity of the training sentences. On \cs{}, we see a clear pattern that more
heads hurt the performance. The \de{} set has more variations to consider but
the results are less conclusive.

For the similarity results, it is worth noting that cs-\textsc{attn-attn} performs very well with 1 attention head but fails miserably with more heads.
Otherwise, the relation to the number of heads is less clear.

\cref{tab:senteval-truncated} compares our strongest models with other approaches on all tasks.
Besides InferSent and GloVe-BOW, we include SkipThought as evaluated by \citet{Conneau2017SupervisedLO}, and the NMT-based
embeddings by \citet{Hill2016LearningDR} trained on the English-French
WMT15 dataset (this is the best result reported by \citeauthor{Hill2016LearningDR} for NMT).

We see that the supervised InferSent clearly outperforms all other models in all tasks
except for MRPC and TREC.
Results by \citeauthor{Hill2016LearningDR} are always lower than our best setups, except MRPC.
On classification tasks, our models are outperformed even by GloVe-BOW, except for the NLI tasks (SICK-E and SNLI) where cs-\textsc{final-ctx} is better.

\subsection{ Paraphrase Scores}
\cref{tab:senteval-para} also provides our measurements based on sentence
paraphrases.
For paraphrase retrieval (NN), we found cosine distance to work better than $L_2$ distance.
We therefore do not list or further consider $L_2$-based results (except in the supplementary material).

This evaluation seems less stable and discerning than the previous
two, but we can again confirm the victory of InferSent followed by our
non-attentive \cs{} models. \cs{} and \de{} models are no longer clearly
separated.

\section{Discussion}
\label{discussion}

\begin{figure}[t]
\centering\scriptsize
\includegraphics[width=\linewidth]{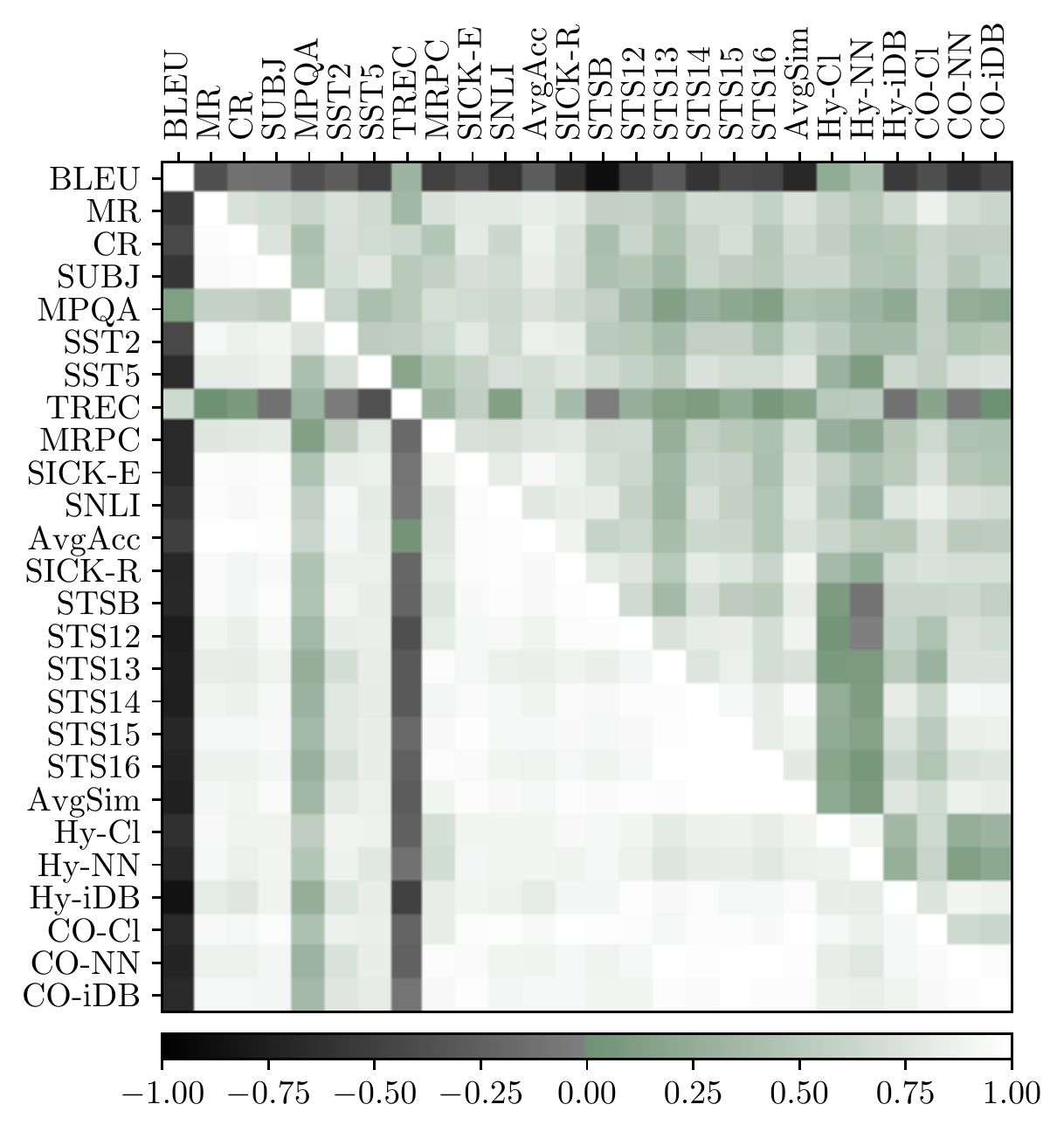}
\caption{Pearson correlations. Upper triangle: \de{} models, lower triangle: \cs{} models. Positive values shown in shades of green. For similarity tasks, only the Pearson (not Spearman) coefficient is represented.}
\label{fig:correlation-heatmap}
\end{figure}

\begin{figure}[t]
\centering\scriptsize
\includegraphics[width=\linewidth]{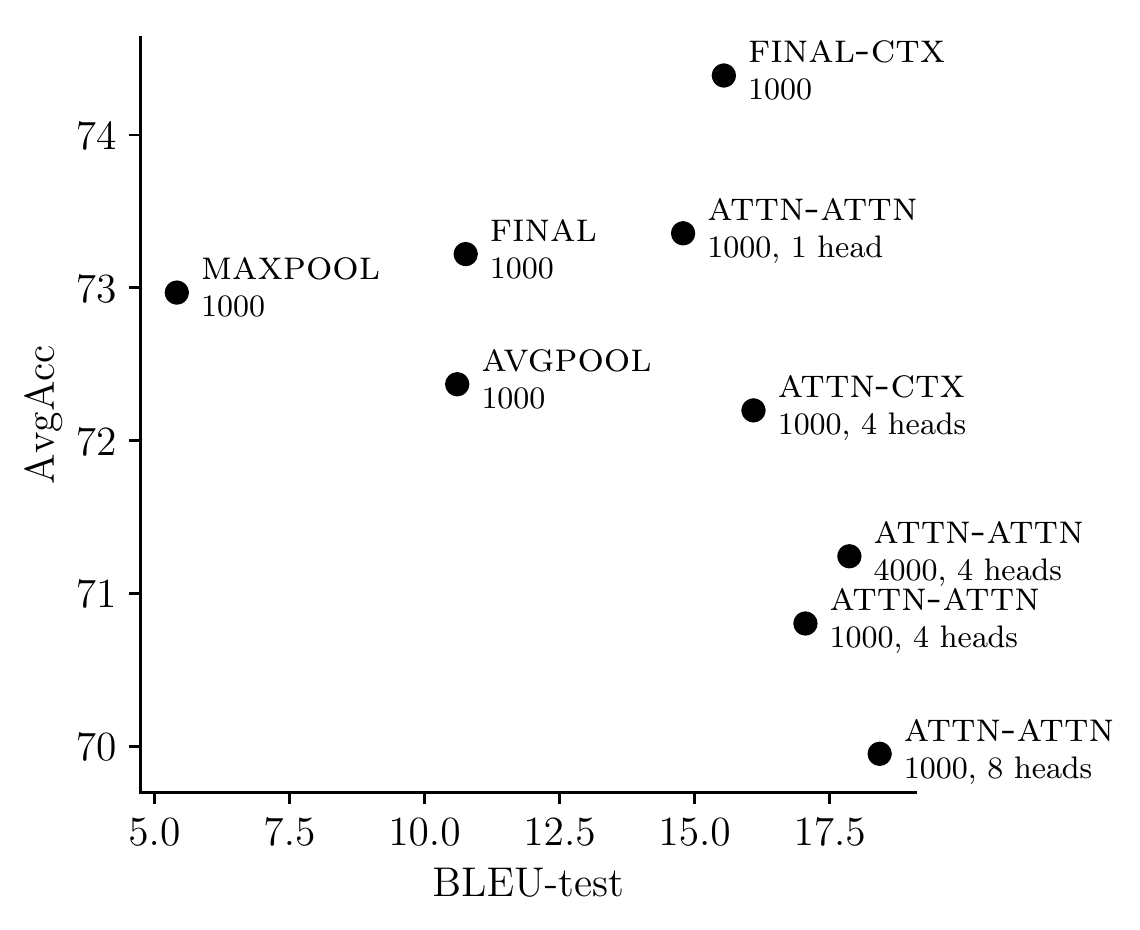}
\caption{BLEU vs. AvgAcc for \cs{} models.}
\label{fig:correlation-bleu-avgacc}
\end{figure}

To assess the relation between the
various measures of sentence representations and translation quality as estimated by BLEU, we plot a heatmap of Pearson correlations
in \cref{fig:correlation-heatmap}. As one example, \cref{fig:correlation-bleu-avgacc} details the \cs{} models' BLEU scores and AvgAcc (average of SentEval accuracies).

\begin{figure}[t]
\centering
\includegraphics[width=\linewidth]{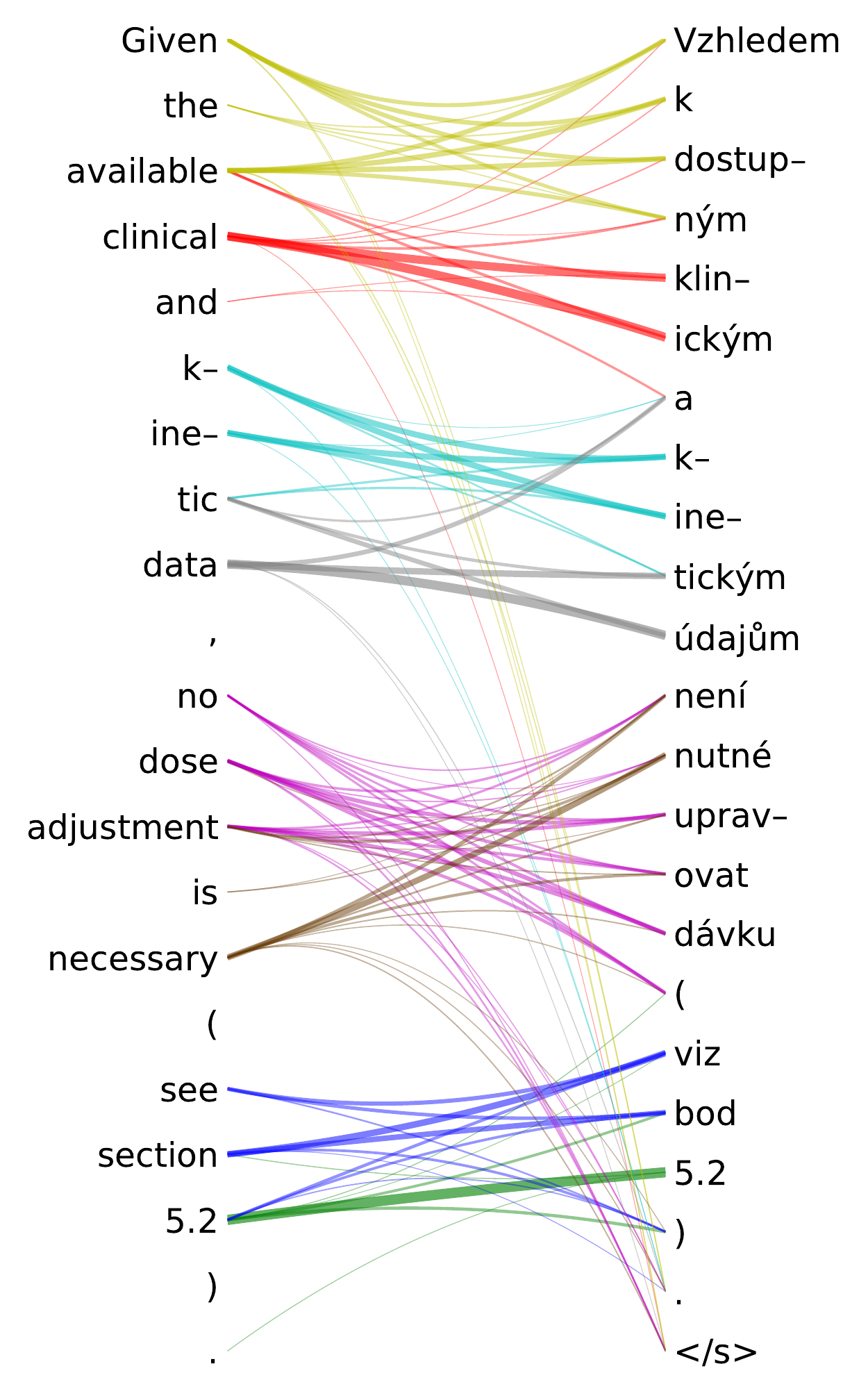}
\caption{Alignment between a source sentence (left) and the output (right) as represented in the \textsc{attn-attn}  model with 8 heads and size of 1000.
Each color represents a different head; the stroke width indicates the alignment weight; weights $\leq 0.01$ pruned out.
(Best viewed in color.)}
\label{fig:align-heads}
\end{figure}

A good sign is that on the \cs{} dataset, most metrics of representation
are positively correlated (the pairwise Pearson correlation is $\csReprCorrelationMean$ on average), the outlier being TREC ($\csReprTRECCorrelationMean$ correlation with the other metrics on average)

On the other hand, most
representation metrics correlate with BLEU negatively ($\csReprBLEUCorrelationMean$) on \cs{}.
The pattern is less pronounced but still clear also on the \de{} dataset.

A detailed understanding of what the learned representations contain is
difficult. We can only speculate that if the NMT model has some capability
for following the source sentence superficially, it will use it and spend its
capacity on closely matching the target  sentences rather than on deriving
some representation of meaning which would reflect e.g.\ semantic similarity. We assume
that this can be a direct consequence
of NMT being trained for cross entropy: putting the exact word forms in exact
positions as the target sentence requires. Performing well in single-reference
BLEU is not an indication that the system understands the meaning but rather
that it can maximize the chance of producing the n-grams required by the
reference.

\begin{figure}[t]
\centering
\includegraphics[width=\linewidth]{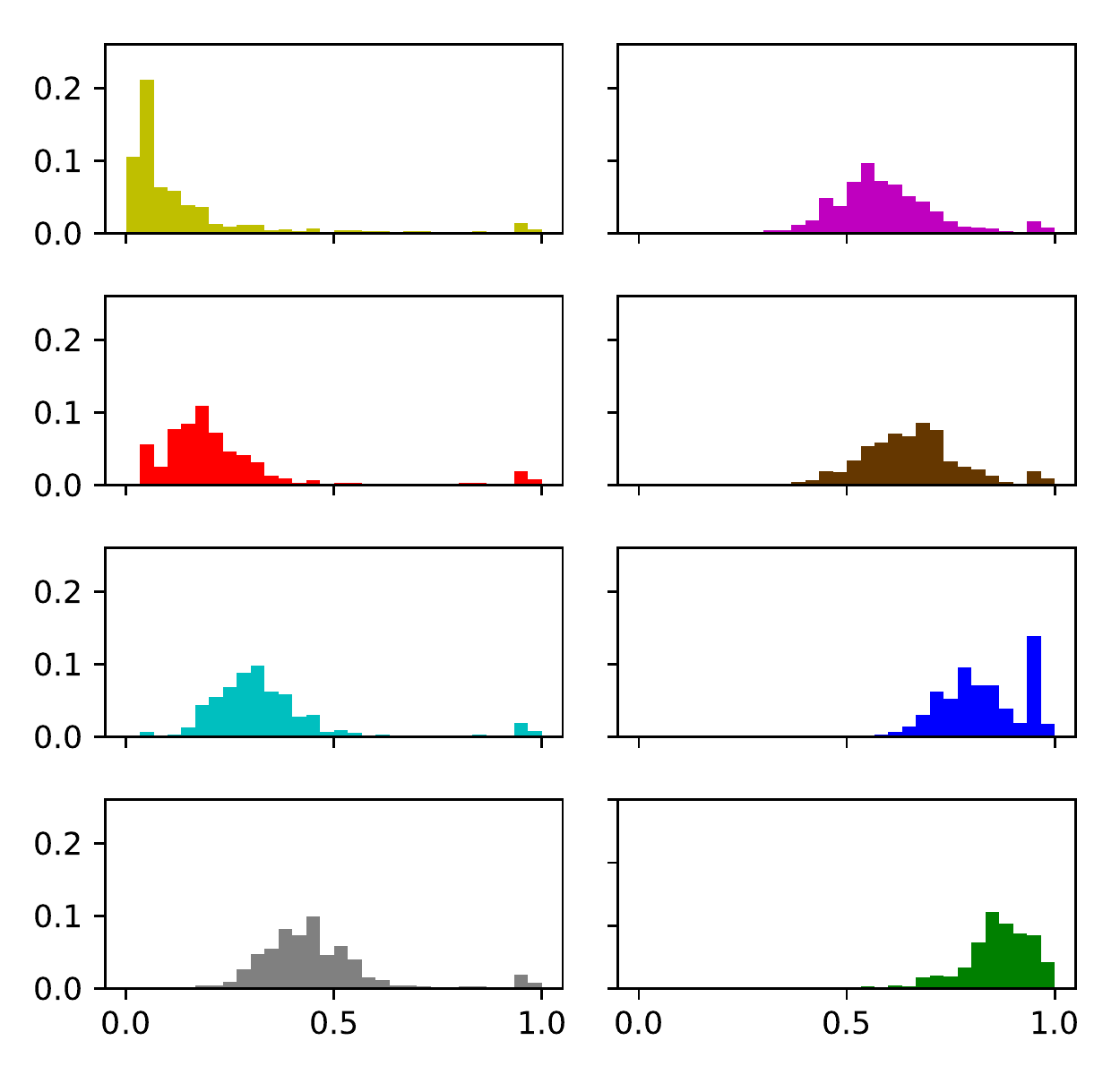}
\caption{Attention weight by relative position in the source sentence (average over dev set excluding sentences shorter than 8 tokens).
Same model as in \cref{fig:align-heads}.
%Each plot corresponds to one of the eight attention heads.
Each plot corresponds to one head.
}
\label{fig:att-hist}
\end{figure}

The negative 
correlation between the number of attention heads and the representation metrics from \cref{fig:correlation-bleu-avgacc} ($\csReprHeadsCorrelationMean$ for \cs{} and $\deReprHeadsCorrelationMean$ for \de{}, on average)
can be
partly explained by the following observation.
We plotted the induced alignments (e.g.\ 
\cref{fig:align-heads}) and noticed that the heads tend to \equo{divide} the
sentence into segments. 
While one would hope that the segments correspond to some meaningful units of
the sentence (e.g.\ subject, predicate, object), we failed to
find any such interpretation for \textsc{attn-attn} and for \cs{} models in general. Instead, the heads divide the source sentence more or
less equidistantly, as documented by \cref{fig:att-hist}. Such a multi-headed
sentence representation is then \emph{less} fit for representing e.g.\ 
paraphrases where the subject and object swap their position due to
passivization, because  their representations are then accessed by different heads, and thus end up in different parts of
the sentence embedding vector.

For de-\textsc{attn-ctx} models, we observed a much flatter distribution of attention weights for each head and, unlike in the other models, we were often able to identify a head focusing on the main verb. This difference between \textsc{attn-attn} and some \textsc{attn-ctx} models could be explained by the fact that in the former, the decoder is oblivious to the  ordering of the heads (because of decoder attention), and hence it may not be useful for a given head to look for a specific syntactic or semantic role.

\section{Conclusion}
\label{conclusion}

We presented a novel variation of attentive NMT models
\parcite{Bahdanau2014NeuralMT,Vaswani2017AttentionIA} that again provides a
single meeting point with a continuous representation of the source sentence.
We evaluated these representations with a number of measures reflecting
how well  the meaning of the source sentence is captured.

While our proposed \equo{compound attention} leads to translation quality not
much worse than the fully attentive model, it generally does not perform well in
the meaning representation. Quite on the contrary, the better the BLEU score,
the worse the meaning representation.

We believe that this observation is important for 
representation learning where bilingual MT now seems less likely to provide
useful
data, but perhaps more so for MT itself, where the struggle towards a high
single-reference BLEU score (or even worse, cross entropy) leads to systems that
refuse to consider the meaning of the sentence.

\section*{Acknowledgement}

This work has been supported by
the grants
18-24210S of the Czech Science Foundation, % GACR Bojar 2018-2020
SVV~260~453
% projekt Specifickeho Vysokoskolskeho Vyzkumu (O. Čepek) 2017-2019
and ``Progress'' Q18+Q48 of Charles University,
and using language resources distributed by
the LINDAT/CLARIN project of the Ministry
of Education, Youth and Sports of the Czech
Republic (projects LM2015071 and OP VVV
VI CZ.02.1.01/0.0/0.0/16 013/0001781).

\bibliography{bibliography}
\bibliographystyle{acl_natbib}

\clearpage{}
\appendix
    \pagenumbering{arabic}
    \setcounter{page}{1}
\onecolumn{}
\section*{Supplementary Material}
%for the Paper: \\\ourtitle{}

\setcounter{table}{0}
\setcounter{figure}{0}
\renewcommand{\thetable}{S\arabic{table}}
\renewcommand\thefigure{S\arabic{figure}}
\renewcommand{\theHtable}{Supplement.\thetable}
\renewcommand{\theHfigure}{Supplement.\thefigure}

\begin{table*}[h]
\centering\fontsize{8.4}{10}\selectfont
\begin{tabular}{l@{~~}r@{~~}r|@{}r@{~~}r@{~~}r@{~~}r@{~~}r@{~~}r@{~~}r@{~~}r@{~~}r@{~~}r|r}
%\toprule
Model &  Size & Heads &    MR &    CR &  SUBJ &  MPQA &  SST2 &  SST5 &  TREC &  MRPC & SICK-E &  SNLI & AvgAcc \\
%\midrule
\hline
\multicolumn{3}{l|}{Most frequent baseline} & 50.0 & 63.8 & 50.0 & 68.8 & 49.9 & 23.1 & 18.8 & 66.5 & 56.7 & 34.3 & 48.2 \\
%\midrule
\hline
\citeauthor{Hill2016LearningDR} en$\rightarrow$fr$^\dagger$  & 2400 & --- & 64.7 & 70.1 & 84.9 & 81.5 & --- & --- & 82.8 & 96.1 & --- & --- & --- \\
SkipThought-LN$^\dagger$ & --- & --- & 79.4 & 83.1 & 93.7 & 89.3 & 82.9 & --- & 88.4 & --- & 79.5 & --- & --- \\
InferSent$^\dagger$ & 4096 & --- & 81.1 & 86.3 & 92.4 & 90.2 & 84.6 & --- & 88.2 & 76.2 & 86.3 & 84.5 & --- \\
\SentEvalA
\hline
\multicolumn{3}{l|}{LM perplexity (\cs{})} & 1362.5 & 736.4 & 1059.0 & 3213.3 & 2099.1 & 1340.8 & 338.2 & 863.0 &          299.4 & 190.6 & 1150.2 \\
\multicolumn{3}{l|}{\% OOV (\cs{})} & 4.2 & 2.5 &  3.6 &  0.9 &  3.4 &  4.2 &  0.6 &  3.5 &            0.2 &  0.3 & 2.3 \\
\multicolumn{3}{l|}{LM perplexity (\de{})} & 3776.8 & 2639.3 & 3137.7 & 8740.0 & 5003.3 & 3519.2 & 3790.8 & 5070.7 &           65.0 & 38.8 & 3578.2 \\
\multicolumn{3}{l|}{\% OOV (\de{})} & 22.8 & 13.1 & 21.0 & 27.9 & 24.4 & 23.3 & 16.7 & 25.6 &            1.7 &  1.5 & 17.8 \\
%\bottomrule
\end{tabular}
\caption{Classification accuracy of sentence representations on a number of SentEval
tasks. Reprinted results are marked with $\dagger$, others are
our measurements.
We give the out-of-vocabulary (OOV) rate and the perplexity  of a 4-gram language model (LM) trained on the English side of the respective parallel corpus and evaluated on all available data for a given task.
\equo{AvgAcc} is the average of each row.
}
%The most frequent baselines for MR, CR, SUBJ and MPQA were estimated on the whole dataset, since there is no train/test split.
\label{tab:acc-full}
\end{table*}

\begin{table*}[t]
\centering\scriptsize
\begin{tabular}{l@{~~}r@{~~}r|r@{~~}r@{~~}r@{~~}r@{~~}r@{~~}r@{~~}r|r}
Model &  Size & Heads &   SICK-R &     STSB &    STS12 &    STS13 &    STS14 &    STS15 &    STS16 & AvgSim\\
\hline
SkipThought-LN$^\dagger$ & --- & --- & \makebox[\widthof{.88/.83}][l]{.85/\,\,---} & --- & --- & --- & .44/.45 & --- & --- & --- \\
InferSent$^\dagger$ & 4096 & --- & \makebox[\widthof{.88/.83}][l]{.88/\,\,---} & --- & --- & --- & .70/.67 & --- & --- & --- \\
\SentEvalB
\hline
\multicolumn{3}{l|}{LM perplexity (\cs{})} & 299.4 & 1338.8 & 697.2 & 2783.9 & 1716.8 & 995.6 & 737.8 & 1224.2 \\
\multicolumn{3}{l|}{\% OOV (\cs{})} & 0.2 & 3.6 &   2.9 &   2.6 &   3.3 &   2.5 &   3.0 & 2.6\\
\multicolumn{3}{l|}{LM perplexity (\de{})} & 65.0 &       1301.4 & 1621.0 & 5041.8 & 2364.6 & 1096.7 & 2583.5 & 2010.6 \\
\multicolumn{3}{l|}{\% OOV (\de{})} & 1.7 & 19.6 &  18.5 &  23.5 &  19.9 &  13.3 &  17.2 & 16.2 \\
\end{tabular}
\caption{Similarity scores of sentence representations: Pearson/Spearman
correlation between cosine similarity of pairs of sentence embeddings and
similarity as marked manually by humans. Reprinted results are marked with $\dagger$, others are
our measurements. \equo{AvgSim} averages both correlation
coefficients for all tasks. Perplexity and OOV rate as in \cref{tab:acc-full}.}
\label{tab:sim-full}
\end{table*}

\begin{table*}[t]
\centering\scriptsize
\begin{tabular}{l@{~~}r@{~~}r|r@{~~~~}r@/r@{~~~~}r|r@{~~~~~~~}r@/r@{~~~~~~~}r|r}
Name &  Size & Heads &  Hy-Cl &   \multicolumn{2}{c}{\hspace{-1em}\clap{Hy-NN ($L_2$/$\cos$)}} & Hy-iDB &  CO-Cl &  \multicolumn{2}{c}{\hspace{-2em}\clap{CO-NN ($L_2$/$\cos$)}} & CO-iDB & AvgPara \\
\hline
\ParaEval
\hline
\multicolumn{3}{l|}{LM perplexity\ \,/\ \,\% OOV (\cs{})} & \multicolumn{4}{c|}{668.5\ \,/\ \,1.2}  & \multicolumn{4}{c|}{238.5\ \,/\ \,0.1} \\
\multicolumn{3}{l|}{LM perplexity\ \,/\ \,\% OOV (\de{})} & \multicolumn{4}{c|}{3354.8\ \,/\ \,19.3}  & \multicolumn{4}{c|}{86.3\ \,/\ \,1.9} \\
\end{tabular}
\caption{Evaluation by paraphrases on data from HyTER (Hy-) and COCO (CO-). \equo{AvgPara}
is simply the average of each row. Perplexity and OOV rate as in \cref{tab:acc-full}.}
\label{tab:para-full}
\end{table*}

\end{document}